\documentclass[conference]{IEEEtran}
\IEEEoverridecommandlockouts
\usepackage{graphicx} 
\usepackage{pifont}
\usepackage{amsmath}
\usepackage{amssymb}  
\usepackage{amsfonts}
\usepackage{algorithmic}
\usepackage{array}
\usepackage{makecell}

\usepackage{multicol}
\usepackage{diagbox}
\usepackage{textcomp}
\usepackage{stfloats}
\usepackage{url}
\usepackage{verbatim}
\usepackage{graphicx}
\usepackage{caption}
\usepackage{subcaption}
\usepackage[ruled, vlined, linesnumbered]{algorithm2e}
\usepackage{cite}
\usepackage{booktabs}
\usepackage{amssymb,amsfonts,bm}
\usepackage{amsmath,tikz}
\usepackage{color}
\usepackage{mathtools}
\usepackage[hidelinks]{hyperref} 
\usepackage{rotating}
\usepackage{blkarray}
\usepackage{physics}
\usetikzlibrary{arrows}
\usepackage{makecell}

\usepackage{amsthm}
\usepackage{comment}
\usepackage{multirow}
\usepackage{geometry}
\graphicspath{{images/}}

\begin{document}
\bstctlcite{IEEEexample:BSTcontrol}

\newcommand{\Dweb}{\mathcal{D}_{\text{web}}}
\newcommand{\Xhat}{\hat{\mathbf{X}}}
\newcommand{\yhat}{\hat{y}}
\newcommand{\Vi}{v_i}
\newcommand{\Zi}{\mathcal{Z}_i}
\newcommand{\Det}{\mathrm{Det}}
\newcommand{\EKF}{\mathrm{EKF}}

\title{ 
Audio-based UAV Localization with Adaptive Temporal Correspondence via Reinforcement Learning
}



\author{Haoxiang Lei, Mingzheng Feng, Daotong Wang, Shenghai Yuan}

\maketitle

\vspace{-2em}
\begin{abstract}
Audio-based localization provides a low-cost and illumination-independent sensing solution for anti-UAV early warning. However, existing methods typically rely on a predefined fixed audio segment length, which limits temporal correspondence and creates a trade-off between sufficient acoustic evidence and timely localization. To address this issue, we propose an audio-based localization framework with adaptive temporal correspondence. A probe segment is first used to extract a compact acoustic state that characterizes the reliability and consistency of the observation. Guided by the state, a reinforcement learning controller dynamically determines the required audio window size for each localization decision. The selected audio segment is then processed by a Mamba-based localization network with adaptive temporal feature modulation for 3D position estimation. Extensive experiments demonstrate that our method achieves competitive 3D localization accuracy with substantially reduced temporal correspondence latency compared to SOTA methods and exhibits strong generalization across scenarios.

\end{abstract}


\section{Introduction}

\begin{figure*}
\centering
\includegraphics[width=16cm]{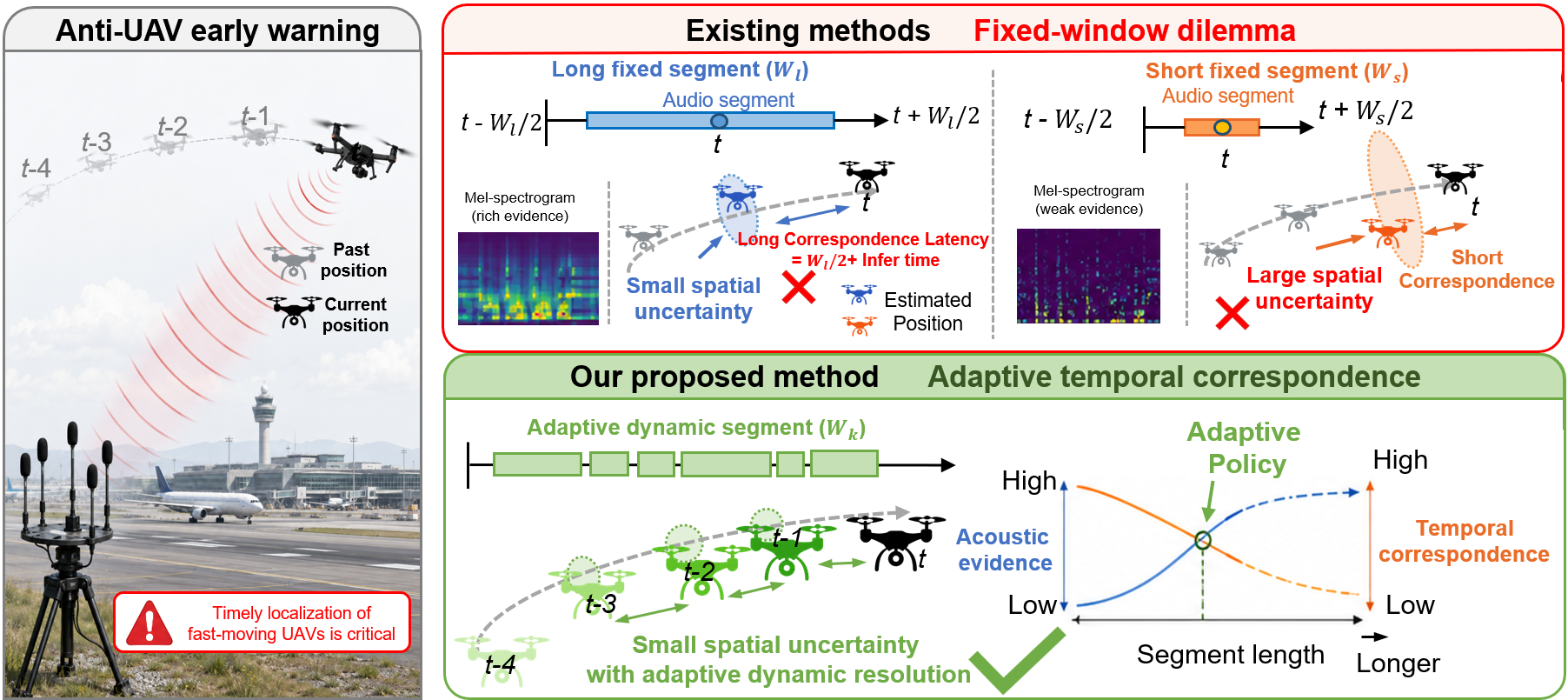}
\caption{Illustration of the fixed-window dilemma in audio-based UAV localization and our adaptive temporal-correspondence framework, which dynamically balances acoustic evidence and temporal correspondence latency for timely anti-UAV early warning.}
\label{fig:motivation}
\end{figure*}



The increasing use of unmanned aerial vehicles (UAVs) in security-sensitive environments has created an urgent demand for effective anti-UAV early-warning systems. Among different sensing modalities \cite{anti-uav1, YANG2024, Det-Fly, LiangUnsupervised, lei2026iros}, audio-based localization \cite{AAUTE, tame, LEI_ESWA} provides low-cost, passive, and illumination-independent sensing, making it particularly attractive for timely UAV localization.





A fundamental requirement of the anti-UAV early warning system is to instantly localize a non-cooperative UAV with both \textbf{high spatial accuracy} and \textbf{high temporal correspondence}. For reliable audio-based early warning, a centered audio window is commonly adopted to balance timely response against sufficiently reliable acoustic evidence, while temporally aligning the acoustic observation with the target state at the window center to reduce motion-induced mismatch \cite{centered-window_Yang2020TellingLF,centered_windowChen21,centeredwindow_mo,centered-window-guo2023dual}. Under this setting, the temporal correspondence is largely determined by the length of the audio segment used for each localization decision. \textbf{However}, \textbf{existing} methods \cite{AAUTE, tame, LEI_ESWA,YANG2024, av-dtec} are biased toward using a predefined fixed segment length, which is often empirically selected and kept unchanged during inference. Long segments accumulate richer acoustic evidence but increase the delay between the represented target state and the available localization result, whereas overly short segments may provide insufficient acoustic cues, as shown in Fig~\ref{fig:motivation}. Therefore, a fixed segment length inherently trades spatial accuracy for temporal correspondence.


Adaptive temporal selection is non-trivial. The \textbf{challenges} lie in: first, the amount of acoustic evidence required for reliable localization varies with target motion, interference, and spatial ambiguity. Second, the window size must be determined online from a short label-free probe before the full observation is accumulated. These coupled factors make hand-crafted adaptive rules difficult to generalize and explain why existing acoustic localization methods typically determine the segment length offline.


\textbf{To address this issue}, we utilize reinforcement learning to adaptively determine how much acoustic observation is necessary for each localization decision. Specifically, we first collect a minimum-duration probe segment and extract a compact acoustic state that characterizes the reliability and consistency of the current observation. Based on the state, an RL-based adaptive controller dynamically selects the appropriate audio window size according to the instantaneous acoustic condition. The selected segment is then transformed into a spectrogram and processed by a Mamba-based localization network for 3D UAV position estimation. In this way, the proposed framework adaptively balances spatial accuracy and temporal correspondence. We conduct extensive experiments on the real-world multimodal anti-UAV dataset MMAUD \cite{MMAUD}, including both in-domain and cross-domain evaluations. The proposed method achieves localization accuracy comparable to state-of-the-art approaches while substantially improving temporal correspondence, enabling the estimated position to more promptly reflect the current state of UAVs. The cross-domain results further demonstrate the robustness and generalization capability of the proposed adaptive localization framework under distribution shifts. Our contributions can be summarized as follows:
\begin{itemize}
\item An audio-based localization framework with adaptive temporal correspondence is proposed for anti-UAV early warning.
\item We design a compact acoustic state to guide RL-based adaptive audio window selection.
\item A Mamba-based audio localization network with adaptive temporal feature modulation is proposed for robust 3D position estimation.
\item Extensive experiments on a real-world dataset demonstrate competitive spatial accuracy, substantially improved temporal correspondence, and strong generalization across scenarios.

\end{itemize}

\section{Related work}

\subsection{Audio-based UAV Localization}

Recent studies have explored acoustic sensing for UAV localization and state estimation. AAUTE \cite{AAUTE} segments continuous audio streams into fixed 2-s clips, converts them into Mel-spectrograms, and employs CNN-based feature extraction for UAV trajectory estimation, followed by Gaussian-process smoothing of the discrete predictions. Lei et al. \cite{LEI_ESWA} divide audio streams into 1-s segments and use an xLSTM-based encoder to extract acoustic features, which are further fused with visual features for UAV 12-DoF state estimation. TAME \cite{tame} processes fixed 1-s audio segments using a Mamba-based network to separately model spectral and temporal information for 3D UAV localization. Similarly, AV-DTEC \cite{av-dtec} adopts 2-s audio segments and asynchronously fuses acoustic and visual observations for UAV trajectory estimation.

Despite their effectiveness, these methods rely on predefined fixed-length audio segments throughout inference. Therefore, dynamically balancing spatial accuracy and temporal correspondence remains an important problem for timely audio-based localization, particularly in real-world anti-UAV early-warning applications.

\subsection{Networks with Dynamic Control}
Dynamic control has recently been introduced into perception networks to adapt their processing strategies to varying sensing conditions. Yuan et al. \cite{yuan2026adaptive} propose A3PRL, where an RL-based controller observes online LiDAR sparsity and tracking statistics and dynamically adjusts voxel resolution, detection sensitivity, and association gating for robust UAV perception. CETUS \cite{cetus} introduces an adaptive-speed controller for event streams, which adjusts the processing step size according to the event rate and latency feedback to balance sampling and inference latency. These studies demonstrate the potential of closed-loop adaptive perception. However, existing dynamic control mainly focuses on spatial perception parameters or inference speed, while adaptive temporal-scale selection for audio-based localization remains largely unexplored.

\section{Proposed Method}
\label{sec:method}

\begin{figure*}
\centering
\includegraphics[width=17cm]{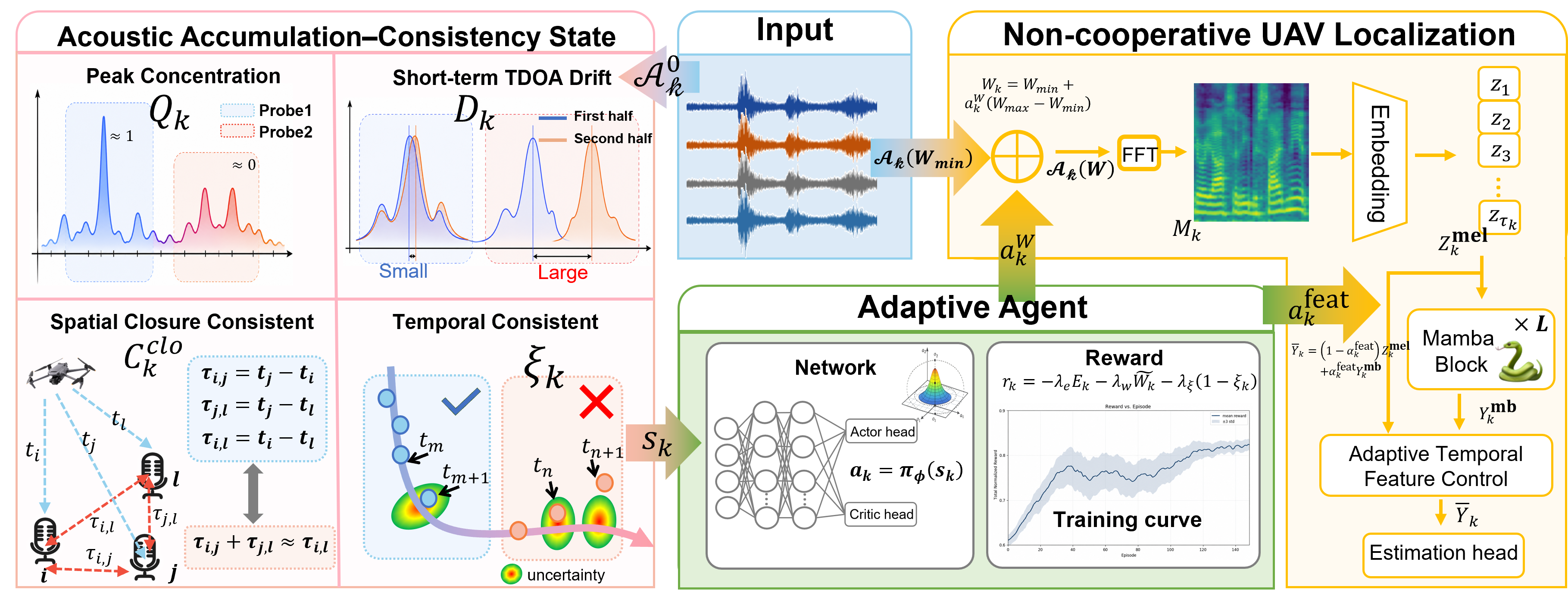}
\caption{Overview of the proposed framework, including the acoustic accumulation-consistency state, adaptive agent with PPO convergence (mean $\pm3$ std over random seeds), and Mamba-based localization backbone.}
\label{fig:network}
\vspace{-1em}
\end{figure*}

As illustrated in Fig.~\ref{fig:network}, we formulate online acoustic localization as a closed-loop adaptive perception problem. Spatial accuracy and temporal correspondence are quantitatively characterized by the \textbf{localization error} and \textbf{temporal correspondence latency}, respectively, with lower values indicating better performance.

\subsection{\textbf{Problem Definition}}
\label{sec:problem}

At localization step \(k\), let
\(\mathcal{A}_k(W)\in\mathbb{R}^{M\times N(W)}\)
denote an \(M\)-channel audio segment of duration
\(W\in\mathbb{R}_{+}\), where
\(N(W)=\lfloor f_sW\rfloor\in\mathbb{N}_{+}\)
and \(f_s\in\mathbb{R}_{+}\) is the sampling frequency.
The objective is to estimate the target position
\(\hat{\mathbf{p}}_k=
[\hat{x}_k,\hat{y}_k,\hat{z}_k]^{\mathsf T}
\in\mathbb{R}^{3}\)
online with low temporal correspondence latency
and stable localization.

Following centered segment--reference alignment
\cite{centered-window_Yang2020TellingLF,centered_windowChen21,centeredwindow_mo,centered-window-guo2023dual}, let $t_k$ denote the reference time at the center of the selected audio window,
i.e., $\mathcal{A}_k(W_k)$ spans $[t_k-W_k/2,\,t_k+W_k/2]$.
The estimate $\hat{\mathbf{p}}_k$ targets the ground-truth state
$\mathbf{p}^{gt}(t_k)$, and therefore the localization error is evaluated at the window-center timestamp. The estimate becomes available at
$t_k^{\mathrm{out}}=t_k+W_k/2+L_i$, yielding the \textbf{temporal correspondence latency} $L_t=t_k^{\mathrm{out}}-t_k=W_k/2+L_i$, where \(L_i\) is the complete inference latency. Thus, localization accuracy and temporal correspondence latency respectively quantify estimation accuracy at the referenced state and its delay to online availability.

To adapt both window length and temporal feature contribution,
we introduce an RL policy \(\pi_\phi\) with observation space
\(\mathcal{S}=[0,1]^4\) and action space
\(\mathcal{U}=[0,1]^2\).
First, a minimum-duration probe
\(\mathcal{A}_k^0=\mathcal{A}_k(W_{\min})\)
is collected, which serves as the central sub-window of the final selected window $\mathcal{A}_k(W_k)$. The extractor \(g(\cdot)\), together with previous tracking feedback, forms the label-free Acoustic Accumulation--Consistency State (AACS):
\(\mathbf{s}_k=[Q_k,D_k,C_k^{\mathrm{clo}}, \bar{\xi}_{k-1}]^{\mathsf T}\in\mathcal{S}\). 

The policy produces
\(\mathbf{a}_k=[a_k^W,a_k^{\mathrm{feat}}]^{\mathsf T}
\in\mathcal{U}\).
The mapping \(h(\cdot)\) converts these actions into
\(W_k=W_{\min}+a_k^W(W_{\max}-W_{\min})\)
and \(\alpha_k^{\mathrm{feat}}=a_k^{\mathrm{feat}}\),
where \(W_{\min}\) and \(W_{\max}\) bound the window length
and \(\alpha_k^{\mathrm{feat}}\) weights temporal features.
The acoustic estimator \(f_\Theta\) then predicts
\(\hat{\mathbf{p}}_k=
f_\Theta(\mathcal{A}_k(W_k);\alpha_k^{\mathrm{feat}})\).

The policy maximizes the expected discounted return,
with reward \(r_k\in\mathbb{R}\) balancing localization
accuracy, window acquisition cost, and tracking consistency.
The closed-loop process is summarized as
\begin{equation}
\mathcal{A}_k^{0}
\xrightarrow{g(\cdot),\,\bar{\xi}_{k-1}}
\mathbf{s}_k
\xrightarrow{\pi_{\phi}}
\mathbf{a}_k
\xrightarrow{h(\cdot)}
(W_k,\alpha_k^{\mathrm{feat}})
\xrightarrow{f_{\Theta}}
\hat{\mathbf{p}}_k.
\end{equation}

\subsection{\textbf{Acoustic Accumulation--Consistency State}}
\label{sec:aacs}

\textbf{Acoustic Accumulation--Consistency State} (AACS) is a compact label-free state:
\begin{equation}
\mathbf{s}_k^{\mathrm{AACS}}
=
\left[
Q_k,
D_k,
C_k^{\mathrm{clo}},
\bar{\xi}_{k-1}
\right]^{\mathsf T}
\in [0,1]^4.
\label{eq:aacs_state}
\end{equation}
The AACS consists of four individual components: GCC Peak Concentration Cue, Short-term TDOA Drift Cue, TDOA Closure-Consistency Cue, and Temporal Consistency Cue.

\subsubsection{\textbf{GCC Peak Concentration Cue}}

Let $\mathcal{P}$ denote the set of microphone pairs \(\mathcal{P} = \left\{ (i,j) \mid 1\leq i<j\leq M \right\}.\) For each pair $(i,j)\in\mathcal{P}$, the GCC-PHAT response computed from the probe segment is defined as
\begin{equation}
G_{ij}^{0}(\tau)
=
\mathcal{F}^{-1}
\left(
\frac{
X_i^{0}(f)X_j^{0*}(f)
}{
\left|
X_i^{0}(f)X_j^{0*}(f)
\right|
+\epsilon
}
\right),
\label{eq:gcc_phat}
\end{equation}
where $X_i^{0}(f)$ and $X_j^{0}(f)$ are the Fourier transforms of the probe signals recorded by microphones $i$ and $j$, respectively.

Given the pairwise microphone distance $d_{ij}\in\mathbb{R}_{+}$ and the speed of sound $c\in\mathbb{R}_{+}$, the delay search is restricted to the physically feasible interval
\begin{equation}
|\tau|
\leq
\tau_{ij}^{\max},
\qquad
\tau_{ij}^{\max}
=
\frac{d_{ij}}{c}.
\label{eq:physical_delay}
\end{equation}
Here, $\tau_{ij}^{\max}\in\mathbb{R}_{+}$ is the maximum physically feasible TDOA magnitude. 
This constraint excludes correlation responses that cannot be produced by acoustic propagation across the corresponding microphone baseline.

Let $\hat{\tau}_{ij}^{0}$ denote the delay associated with the strongest local maximum within the feasible interval:
\begin{equation}
\hat{\tau}_{ij}^{0}
=
\arg\max_{
|\tau|\leq\tau_{ij}^{\max}
}
\left|
G_{ij}^{0}(\tau)
\right|.
\label{eq:dominant_delay}
\end{equation}
Directly selecting the two largest correlation samples may incorrectly treat adjacent samples from the same broad correlation lobe as two independent peaks. We therefore suppress a local neighborhood $\mathcal{N}(\hat{\tau}_{ij}^{0})$ around the dominant peak before searching for the secondary peak. The dominant and secondary peak magnitudes are defined as
\begin{equation}
g_{ij}^{(1)}
=
\left|
G_{ij}^{0}
\left(
\hat{\tau}_{ij}^{0}
\right)
\right|,
\label{eq:primary_peak}
\end{equation}
and
\begin{equation}
g_{ij}^{(2)}
=
\max_{
\substack{
|\tau|\leq\tau_{ij}^{\max},\\
\tau\notin\mathcal{N}(\hat{\tau}_{ij}^{0})
}
}
\left|
G_{ij}^{0}(\tau)
\right|,
\label{eq:secondary_peak}
\end{equation}
where peak magnitudes satisfy $g_{ij}^{(1)},g_{ij}^{(2)}\in\mathbb{R}_{\geq 0}$.
The pairwise peak concentration is then computed as
\begin{equation}
q_{ij}
=
\operatorname{clip}
\left(
\frac{
g_{ij}^{(1)}-g_{ij}^{(2)}
}{
g_{ij}^{(1)}+\epsilon
},
0,1
\right).
\label{eq:pair_peak_concentration}
\end{equation}
A large $q_{ij}$ indicates that the dominant peak is clearly separated from alternative delay hypotheses. The array-level peak concentration is obtained by averaging over all microphone pairs:
\begin{equation}
Q_k
=
\frac{1}{|\mathcal{P}|}
\sum_{(i,j)\in\mathcal{P}}
q_{ij},\quad Q_k\in[0,1].
\label{eq:global_peak_concentration}
\end{equation}

A large $Q_k$ indicates distinctive pairwise TDOA observations, whereas a small value indicates that the current probe segment contains ambiguous correlation peaks.

\vspace{1em}
\subsubsection{\textbf{Short-term TDOA Drift Cue}}



We divide the probe segment into two equal subsegments without overlapping:
\(\mathcal{A}_k^{0} = \mathcal{A}_k^{0,1} \cup \mathcal{A}_k^{0,2}.\) The pairwise delays estimated from the two subsegments are denoted by $\hat{\tau}_{ij}^{0,1}$ and $\hat{\tau}_{ij}^{0,2}$, respectively. The normalized drift of microphone pair $(i,j)$ is defined as
\begin{equation}
d_{ij}^{0}
=
\operatorname{clip}
\left(
\frac{
\left|
\hat{\tau}_{ij}^{0,2}
-
\hat{\tau}_{ij}^{0,1}
\right|
}{
2\tau_{ij}^{\max}+\epsilon
},
0,1
\right), \quad d_{ij}^{0}\in[0,1], 
\label{eq:pair_tdoa_drift}
\end{equation}
where the denominator $2\tau_{ij}^{\max}$ represents the maximum possible delay variation within the physically feasible pairwise range.

To reduce the influence of unstable delay estimates, we compute $q_{ij}^{0,1},q_{ij}^{0,2}\in[0,1]$ for the two subsegments using Eq.~\eqref{eq:pair_peak_concentration}, respectively, and define the pairwise confidence weight as
\begin{equation}
w_{ij}^{0}
=
\sqrt{q_{ij}^{0,1}q_{ij}^{0,2}}
\in[0,1].
\end{equation}
The array-level TDOA drift is then defined as
\begin{equation}
D_k
=
\frac{
\sum_{(i,j)\in\mathcal{P}}
w_{ij}^{0}d_{ij}^{0}
}{
\sum_{(i,j)\in\mathcal{P}}
w_{ij}^{0}
+
\epsilon
},\quad D_k\in[0,1]
\label{eq:tdoa_drift}
\end{equation} 

A small $D_k$ indicates that the spatial delay pattern remains stable within the probe segment, while a large value suggests rapid source motion or temporally unstable TDOA estimation.


\vspace{1em}
\subsubsection{\textbf{Spatial Closure Consistency}}
It evaluates whether the estimated pairwise delays constitute a physically self-consistent spatial observation. We adopt the directed delay convention \(\hat{\tau}_{ij}^{0} = t_j-t_i,\) where $t_i$ and $t_j$ denote the acoustic arrival times at microphones $i$ and $j$, respectively. 
For any microphone triplet $(i,j,l)$, pairwise delays generated by a common acoustic source should satisfy
\begin{equation}
\hat{\tau}_{ij}^{0}
+
\hat{\tau}_{jl}^{0}
-
\hat{\tau}_{il}^{0}
\approx
0.
\label{eq:closure_relation}
\end{equation}
Let $\mathcal{T}$ denote the set of microphone triplets \(\mathcal{T} = \left\{ (i,j,l) \mid 1\leq i<j<l\leq M \right\}.\) For each triplet, the normalized closure residual is defined as:
\begin{equation}
r_{ijl}^{\mathrm{clo}}
=
\frac{
\left|
\hat{\tau}_{ij}^{0}
+
\hat{\tau}_{jl}^{0}
-
\hat{\tau}_{il}^{0}
\right|
}{
\tau_{ij}^{\max}
+
\tau_{jl}^{\max}
+
\tau_{il}^{\max}
+
\epsilon
}, \quad r_{ijl}^{\mathrm{clo}}\in[0,1]
\label{eq:closure_residual}
\end{equation}
The array-level closure residual is obtained as
\begin{equation}
R_k^{\mathrm{clo}}
=
\frac{1}{|\mathcal{T}|}
\sum_{(i,j,l)\in\mathcal{T}}
r_{ijl}^{\mathrm{clo}}, \quad R_k^{\mathrm{clo}}\in[0,1]
\label{eq:global_closure_residual}
\end{equation}
It is converted into a bounded closure-consistency score:
\begin{equation}
C_k^{\mathrm{clo}}
=
\exp
\left(
-\eta_cR_k^{\mathrm{clo}}
\right),
\label{eq:closure_consistency}
\end{equation}
where $\eta_c\in\mathbb{R}_{+}$ determines the sensitivity to spatial inconsistency. Consequently, $C_k^{\mathrm{clo}}\in(0,1]\subset[0,1]$.

A high \(C_k^{\mathrm{clo}}\) indicates that the pairwise TDOAs satisfy the microphone-array closure relation and therefore form a physically self-consistent spatial delay pattern.

\vspace{1em}
\subsubsection{\textbf{Temporal Consistency Cue}}
\label{sec:tracking_consistency}
We introduce a label-free temporal tracking-consistency cue that feeds the downstream localization behavior back to the adaptive controller.

We maintain a target state
$\mathbf{x}_k=
[\mathbf{p}_k^{\mathsf T},
\mathbf{v}_k^{\mathsf T}]^{\mathsf T}
\in\mathbb{R}^{6}$, where $\mathbf{p}_k,\mathbf{v}_k\in\mathbb{R}^{3}$ denote the tracked
3D position and velocity, respectively.
And we adopt a constant-velocity Kalman filter. Given the interval $\Delta t_k\in\mathbb{R}_{+}$ between two localization steps, the predicted state and covariance are
\begin{equation}
\begin{aligned}
\hat{\mathbf{x}}_{k|k-1}
&=
\mathbf{F}_k
\hat{\mathbf{x}}_{k-1|k-1},\\
\mathbf{P}_{k|k-1}
&=
\mathbf{F}_k
\mathbf{P}_{k-1|k-1}
\mathbf{F}_k^{\mathsf T}
+
\mathbf{Q}_k^{\mathrm{proc}},
\end{aligned}
\mathbf{F}_k
=
\begin{bmatrix}
\mathbf{I}_3 & \Delta t_k\mathbf{I}_3\\
\mathbf{0}_3 & \mathbf{I}_3
\end{bmatrix},
\label{eq:kf_prediction}
\end{equation}
where $\mathbf{F}_k\in\mathbb{R}^{6\times6}$ is the state-transition matrix, $\mathbf{P}_{k|k-1}\in\mathbb{R}^{6\times6}$ is the predicted state covariance, and $\mathbf{Q}_k^{\mathrm{proc}}\in\mathbb{R}^{6\times6}$ denotes the process-noise covariance. Here,
$\mathbf{I}_3,\mathbf{0}_3\in\mathbb{R}^{3\times3}$ denote the identity and zero matrices, respectively.

The current audio-derived position estimate
$\hat{\mathbf{p}}_k$ is treated as the position measurement. Its
innovation and innovation covariance are
\begin{equation}
\boldsymbol{\nu}_k
=
\hat{\mathbf{p}}_k
-
\mathbf{H}
\hat{\mathbf{x}}_{k|k-1},
\qquad
\mathbf{S}_k
=
\mathbf{H}
\mathbf{P}_{k|k-1}
\mathbf{H}^{\mathsf T}
+
\mathbf{R}_k,
\label{eq:kf_innovation}
\end{equation}
where $\mathbf{H}=[\mathbf{I}_3\ \mathbf{0}_3]\in\mathbb{R}^{3\times6}$ is the position-observation matrix, $\boldsymbol{\nu}_k\in\mathbb{R}^{3}$ is the innovation vector, $\mathbf{S}_k\in\mathbb{R}^{3\times3}$ is the innovation covariance, and $\mathbf{R}_k=\sigma_r^2\mathbf{I}_3\in\mathbb{R}^{3\times3}$ is the fixed measurement covariance.

The temporal tracking-consistency cue is defined from the normalized
innovation as
\begin{equation}
\xi_k
=
\exp
\left(
-\frac{1}{2}
\boldsymbol{\nu}_k^{\mathsf T}
\mathbf{S}_k^{-1}
\boldsymbol{\nu}_k
\right), \quad \xi_k\in(0,1]\subset[0,1]
\label{eq:tracking_consistency}
\end{equation}
A high $\xi_k$ indicates that the audio-derived localization result is
consistent with the predicted target motion under the current
uncertainty, whereas a low value indicates an abrupt or unreliable
localization update. After evaluating $\xi_k$, the standard Kalman
correction is applied to update the maintained state for the next
localization step.

To reduce short-term fluctuations, we further maintain an
exponentially smoothed consistency cue:
\begin{equation}
\bar{\xi}_k
=
\beta_{\xi}\bar{\xi}_{k-1}
+
\left(
1-\beta_{\xi}
\right)
\xi_k,
\label{eq:consistency_ema}
\end{equation}
where $\beta_{\xi}\in[0,1)$ is the smoothing coefficient.

\subsection{\textbf{PPO-Based Adaptive Policy}}
\label{sec:ppo}


The PPO controller uses the AACS observation to select the audio window and temporal feature weight. Its observation combines three probe-derived acoustic cues with the previous smoothed tracking-consistency cue $\bar{\xi}_{k-1}$, as \(\mathbf{s}_k
=
\left[
Q_k,
D_k,
C_k^{\mathrm{clo}},
\bar{\xi}_{k-1}
\right]^{\mathsf T}
\in[0,1]^4.\)

\vspace{1em}
\noindent\textbf{Actions.}
Given $\mathbf{s}_k$, the policy outputs a two-dimensional continuous
action:
\begin{equation}
\mathbf{a}_k
=
\pi_{\phi}(\mathbf{s}_k)
=
\left[
a_k^{W},
a_k^{\mathrm{feat}}
\right]^{\mathsf T},
\qquad
\mathbf{a}_k\in[0,1]^2.
\label{eq:policy_action}
\end{equation}
The actions are mapped to the final audio accumulation length and the temporal feature weight:
\begin{equation}
W_k
=
W_{\min}
+
a_k^{W}
\left(
W_{\max}-W_{\min}
\right),
\quad
\alpha_k^{\mathrm{feat}} = a_k^{\mathrm{feat}}.
\label{eq:action_mapping}
\end{equation}
The policy therefore learns how the current acoustic condition and tracking stability should jointly regulate the observation length and the use of long-range temporal features.

\vspace{1em}
\noindent\textbf{Reward and optimization.}
The localization error and normalized acquisition cost are defined as
\begin{equation}
E_k
=
\left\|
\hat{\mathbf{p}}_k
-
\mathbf{p}_k^{\mathrm{gt}}
\right\|_2,
\qquad
\widetilde{W}_k
=
\frac{
W_k-W_{\min}
}{
W_{\max}-W_{\min}
}, 
\label{eq:accuracy_latency_terms}
\end{equation} 
where $\mathbf{p}_k^{\mathrm{gt}}\in\mathbb{R}^{3}$ represent the ground truth position, $E_k\in\mathbb{R}_{\geq0}$ and
$\widetilde{W}_k\in[0,1]$.

Together with the temporal tracking-consistency cue defined in
Eq.~\eqref{eq:tracking_consistency}, the reward is
\begin{equation}
r_k
=
-\lambda_eE_k
-\lambda_w\widetilde{W}_k
-\lambda_{\xi}
\left(
1-\xi_k
\right),
\label{eq:reward}
\end{equation}
where $\lambda_e,\lambda_w,\lambda_{\xi}\in\mathbb{R}_{+}$ are weighting coefficients balancing localization accuracy, audio accumulation cost, and motion-consistent trajectory estimation, respectively, and $r_k\in\mathbb{R}$ denotes the scalar reward.

The policy is trained using PPO to maximize the expected discounted
return:
\begin{equation}
\phi^{*}
=
\arg\max_{\phi}
\mathbb{E}_{\pi_{\phi}}
\left[
\sum_{k=0}^{K-1}
\gamma^{k}r_k
\right].
\label{eq:ppo_objective}
\end{equation}
Here, $\gamma\in[0,1)$ denotes the discount factor and $K\in\mathbb{N}_{+}$ is the rollout horizon.
Standard PPO clipping, value-function regression, reward normalization, and entropy regularization are adopted for stable optimization.

\subsection{\textbf{Mamba-Based Acoustic Localization Backbone}}
\label{sec:backbone}
Given the audio window selected by the policy, we compute the multichannel Mel-spectrogram
\begin{equation}
\mathbf{M}_k
=
\mathrm{Mel}\!\left(\mathcal{A}_k(W_k)\right)
\in
\mathbb{R}^{M\times F\times T_k},
\end{equation}
where $M$, $F$, and $T_k$ denote the number of microphones, Mel-frequency bins, and temporal frames, respectively. The spectrogram is then projected into a sequence of $d$-dimensional acoustic tokens:
\begin{equation}
\mathbf{Z}^{\mathrm{mel}}_k
=
\mathrm{Embed}(\mathbf{M}_k)
\in
\mathbb{R}^{T_k\times d}.
\end{equation}

We employ $L$ residual Mamba blocks to model temporal dependencies:
\begin{equation}
\mathbf{Y}^{(\ell+1)}_k
=
\mathbf{Y}^{(\ell)}_k
+
\mathrm{Dropout}\!\left(
\mathrm{Mamba}_{\ell}
\left(
\mathrm{LN}(\mathbf{Y}^{(\ell)}_k)
\right)
\right),
\end{equation}
where $\mathbf{Y}^{(1)}_k=\mathbf{Z}^{\mathrm{mel}}_k$, $\ell=1,\ldots,L$, and $\mathbf{Y}^{(\ell)}_k\in\mathbb{R}^{T_k\times d}$. We denote the final Mamba representation by $\mathbf{Y}^{\mathrm{mb}}_k=\mathbf{Y}^{(L+1)}_k$.

\noindent\textbf{Adaptive Temporal Feature Control:}
The second policy action controls the contribution of the Mamba-enhanced temporal representation:
\begin{equation}
\mathbf{Y}_k
=
(1-\alpha_k^{\mathrm{feat}})
\mathbf{Z}^{\mathrm{mel}}_k
+
\alpha_k^{\mathrm{feat}}
\mathbf{Y}^{\mathrm{mb}}_k
\in
\mathbb{R}^{T_k\times d}.
\end{equation}
A small $\alpha_k^{\mathrm{feat}}$ preserves more local spectrogram information, whereas a large value emphasizes the temporal dependencies modeled by Mamba.

Finally, the adaptive representation is pooled and mapped to the 3D target position:
\begin{equation}
\hat{\mathbf{p}}_k
=
\mathrm{MLP}_{\mathrm{loc}}
\left(
\mathrm{Pool}
\left(
\mathrm{LN}(\mathbf{Y}_k)
\right)
\right)
\in
\mathbb{R}^{3}.
\end{equation}
Thus, $a_k^{W}$ controls the amount of acoustic observation, while $\alpha_k^{\mathrm{feat}}$ controls the contribution of temporal Mamba features.

\begin{table*}[!t]
\footnotesize
\renewcommand*{\arraystretch}{1.1}
\centering
\caption{Comparison on MMAUD-V1 [14].
$D_x,D_y,D_z$: axis-wise MAE;
$E$: mean 3D Euclidean error;
$P_{95}$: 95th-percentile 3D error;
$SR_\tau$: percentage of estimates within $\tau$ meters.
$L_w,L_i,L_t$: window length, total inference latency,
and temporal correspondence latency.
Errors are in meters; times are in seconds.
Best and second-best results are \textbf{bold} and \underline{underlined},
respectively, unless otherwise stated.}
\small
\renewcommand{\arraystretch}{1.3}
\setlength{\tabcolsep}{6.2pt}
\begin{tabular}{lcccccccccccc}
\hline
\toprule
\textbf{Methods} 
& \textbf{Year} 
& \textbf{Modal} 
& \multicolumn{7}{c}{\textbf{Localization Performance on V1}}
& \multicolumn{3}{c}{\textbf{Latency}}
\\
\cmidrule(lr){4-10} 
\cmidrule(lr){11-13}

&
&
&
$D_x\downarrow$
& $D_y\downarrow$
& $D_z\downarrow$
& $E\downarrow$
& $P_{95}\downarrow$
& $SR_{0.3}\uparrow$
& $SR_{0.5}\uparrow$
& $\mathcal{L}_w$
& $\mathcal{L}_i$
& $\mathcal{L}_t$
\\
\midrule

\textbf{AudioNet} \cite{yang2023av}
& 2023 & Audio-Only
& 0.19 & 0.43 & 0.46 & 0.76
& 1.72 & 18.6 & 35.8
& 2.0 & 0.023 & 1.023
\\

\textbf{VorasNet} \cite{vora2023dronechase}
& 2023 & Audio-Only
& 0.28 & 0.61 & 0.61 & 1.06
& 2.31 & 12.3 & 25.4
& 1.0 & 0.021 & \underline{0.521}
\\

\textbf{TAME} \cite{tame}
& 2025 & Audio-Only
& \textbf{0.11} & 0.30 & 0.34 & 0.55
& 1.24 & 30.7 & 49.6
& 1.0 & 0.024 & 0.524
\\

\textbf{AAUTE} \cite{AAUTE}
& 2025 & Audio-Only
& 0.14 & 0.26 & 0.25 & 0.48
& 1.08 & 36.5 & 56.8
& 2.0 & \textbf{0.018} & 1.018
\\

\hline

\textbf{ASDNet} \cite{tao2021someone}
& 2021 & Audio+Image
& 0.31 & 0.69 & 0.44 & 0.99
& 2.12 & 13.7 & 27.9
& 1.0 & 0.053 & 0.553
\\

\textbf{AV-PED} \cite{yang2023av}
& 2023 & Audio+Image
& 0.31 & 0.43 & 0.52 & 0.87
& 1.89 & 16.2 & 31.7
& 1.0 & 0.047 & 0.547
\\

\textbf{AV-FDTI} \cite{YANG2024}
& 2024 & Audio+Image
& \underline{0.13} & 0.23 & 0.36 & 0.53
& 1.17 & 33.4 & 52.3
& 1.0 & 0.052 & 0.552
\\

\textbf{Lei et al.} \cite{LEI_ESWA}
& 2026 & Audio+Image
& 0.15 & \textbf{0.16} & \underline{0.19} & \textbf{0.29}
& \underline{0.69} & \underline{68.7} & \underline{87.4}
& 1.0 & 0.095 & 0.595
\\

\textbf{AV-DTEC} \cite{av-dtec}
& 2026 & Audio+Image
& 0.33 & 0.25 & 0.27 & 0.58
& 1.29 & 27.8 & 46.1
& 2.0 & 0.086 & 1.086
\\

\hline

\textbf{Ours}
& -- & Audio-Only
& 0.17 & \underline{0.18} & \textbf{0.12} & \underline{0.30}
& \textbf{0.61} & \textbf{72.6} & \textbf{89.2}
& \textbf{0.23} & \underline{0.021} & \textbf{0.136}
\\

\bottomrule
\end{tabular}
\label{tab:comparison_mmaud1}
\end{table*}





\subsection{\textbf{Training and Inference}}

Training comprises three stages.
Stage I pretrains the acoustic estimator with
$W_k=W_{\max}$ and $\alpha_k^{\mathrm{feat}}=1$.
Stage II trains it under randomized controls:
\(W_k\sim\mathcal{U}(W_{\min},W_{\max}),
\alpha_k^{\mathrm{feat}}\sim\mathcal{U}(0,1).\) Both stages minimize the mean Euclidean localization loss:
\begin{equation}
\mathcal{L}_{\mathrm{loc}}(\Theta)
=
\frac{1}{N}\sum_{k=1}^{N}
\left\|\hat{\mathbf{p}}_k-\mathbf{p}_k^{\mathrm{gt}}\right\|_2.
\end{equation}
Stage III freezes the estimator and optimizes only
the PPO controller.
Ground truth supervises Stages I--II and provides
the reward in Stage III; inference requires no labels.

During inference, the system computes AACS from the probe
and previous tracking feedback, selects
$(W_k,\alpha_k^{\mathrm{feat}})$,
and localizes the target using the selected audio window.

\vspace{-1em}
\section{Experiments and Evaluation}

\subsection{Experimental Setup}

\begin{figure}
\centering
\includegraphics[width=3.2in]{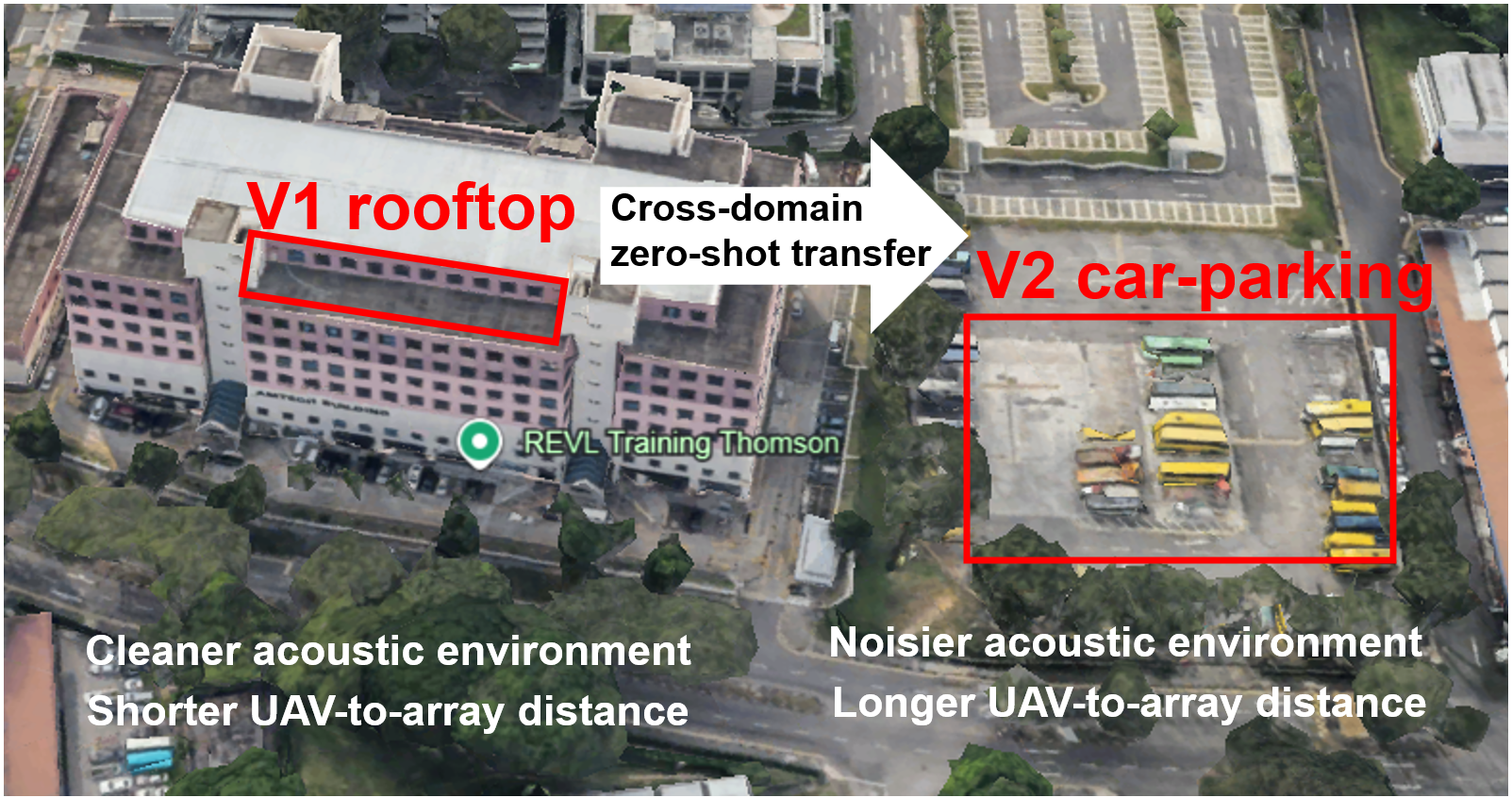}
\caption{Real-world MMAUD acquisition scenarios for cross-domain evaluation, where V2 presents stronger acoustic noise and more distant UAV targets than V1.}
\label{fig:cross}
\end{figure}

Experiments are conducted on the real-world multimodal MMAUD dataset \cite{MMAUD}, which provides audio, image, radar, LiDAR, and Leica MS60 ground-truth 3D trajectories. We use the rooftop scene of MMAUD-V1 for in-domain evaluation and the car-parking scene of MMAUD-V2 for zero-shot cross-domain evaluation. Training and testing sets are split at the complete-sequence level, and no temporally overlapping audio segments are shared across different subsets.

The PPO actor and critic are two-layer MLPs with 64 hidden units and ReLU activations, with the actor parameterizing a sigmoid-bounded diagonal Gaussian policy. PPO is optimized using Adam with a learning rate of $3\times10^{-4}$, a clipping coefficient $\epsilon_{\mathrm{ppo}}=0.2$, a discount factor $\gamma=0.99$, and a GAE coefficient of $0.95$. The entropy and value-loss coefficients are set to $0.01$ and $0.5$, respectively. Each policy update uses 2048 transitions, a mini-batch size of 64, and 10 optimization epochs, with the gradient norm clipped to $0.5$. We set $W_{\min}=0.1$~s and $W_{\max}=2.0$~s. The reward weights are set to $\lambda_e=1.0$, $\lambda_w=0.2$, and $\lambda_{\xi}=0.1$, while the temporal smoothing coefficient is $\beta_{\xi}=0.8$. For temporal-consistency estimation, we set $\sigma_r=0.5$~m for the measurement noise. All experiments are conducted on a workstation equipped with a single NVIDIA GeForce RTX 4090 GPU.

During inference, consecutive localization windows may overlap, with shared raw audio retained in the streaming buffer. For conservative latency evaluation, we recompute the selected window at each step without cross-window feature caching.

\subsection{Performances}

\begin{table}[h]
\centering
\caption{\textbf{Cross-domain} robustness evaluation on the MMAUD v2~\cite{MMAUD}.}
\label{table:comparisonmmaud2_cross}
\renewcommand{\arraystretch}{1.2}
\setlength{\tabcolsep}{3.5pt}
\small
\begin{tabular}{lcccccccc}
\toprule
\textbf{Methods} 
& \multicolumn{4}{c}{\textbf{Position Accuracy on V2}} 
& \multicolumn{3}{c}{\textbf{Latency}} \\ 
\cmidrule(lr){2-5} 
\cmidrule(lr){6-8} 
& $D_x\downarrow$ 
& $D_y\downarrow$ 
& $D_z\downarrow$ 
& \textbf{$E\downarrow$} 
& $\mathcal{L}_w$ 
& $\mathcal{L}_i$
& $\mathcal{L}_t$ \\ 
\hline
\textbf{AudioNet} \cite{yang2023av} & 1.53 & 0.94 & 1.26 & 2.89 & 2.0 & 0.023 & 1.023\\
\textbf{VorasNet} \cite{vora2023dronechase} & 1.24 & 1.31 & 1.23 & 3.57 &  1.0  & \underline{0.021}  & \underline{0.521}\\
\textbf{TAME} \cite{tame}&  1.43 & 0.78 & 1.04 & 2.34 & 1.0 & 0.024 & 0.524\\
\textbf{AAUTE} \cite{AAUTE}& 1.39 & 1.05 & 0.98 & 2.56 & 2.0 & \textbf{0.018} & 1.018\\
\hline
\textbf{ASDNet} \cite{tao2021someone}& 0.97 & 1.23 & 1.02 & 2.03 & 1.0 & 0.053 & 0.553\\
\textbf{AV-PED} \cite{yang2023av}& 1.03 & 1.63 & 1.23 & 2.65 & 1.0 & 0.047 & 0.547 \\
\textbf{AV-FDTI} \cite{YANG2024} & \underline{0.78} & 0.89 & 1.04 & 1.78 & 1.0  & 0.052 & 0.552\\
\textbf{Lei et al.} \cite{LEI_ESWA} & 0.54 & \textbf{0.58} & \underline{0.79} & \underline{1.32} & 1.0 & 0.095 & 0.595\\
\textbf{AV-DTEC} \cite{av-dtec} & 1.03 & 0.94 & 1.06 & 2.21 & 2.0 & 0.086 & 1.086\\
\hline
\textbf{Ours}&  \textbf{0.48} & \underline{0.65} & \textbf{0.45} & \textbf{0.96} & \textbf{0.72} & 0.029 & \textbf{0.389} \\
\bottomrule
\end{tabular}
\end{table}

\begin{figure}
\centering
\includegraphics[width=2.8in]{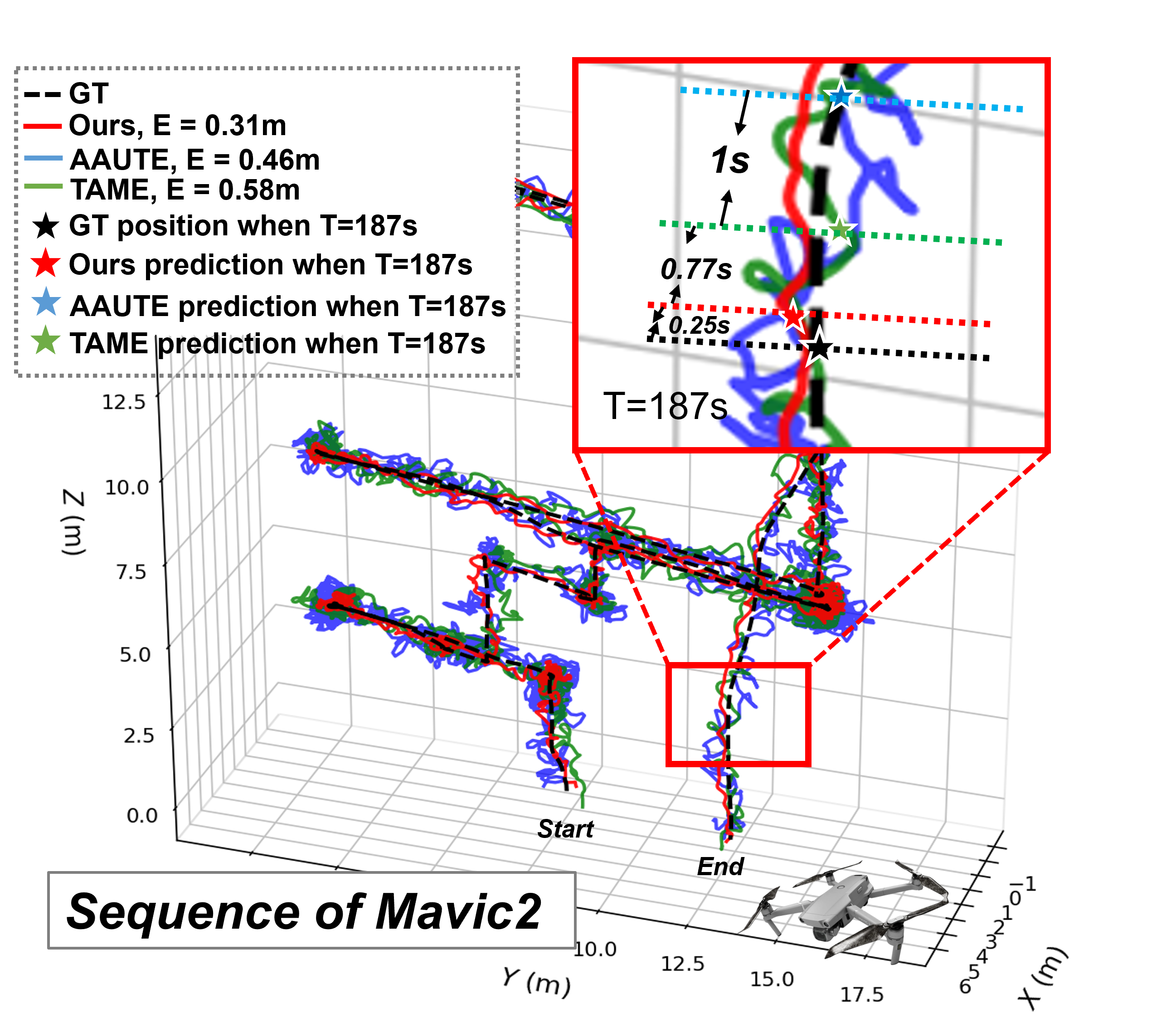}
\caption{Trajectory comparison on the MMAUD-V1 Mavic2 sequence. The enlarged view shows the localization offsets at $T=187$~s caused by different audio accumulation latencies.}
\label{fig:traj_comparison}
\label{fig:traj_comparison}
\end{figure}

\textbf{On MMAUD \cite{MMAUD}: }
Table~\ref{tab:comparison_mmaud1} compares our method with representative audio-only and audio-visual baselines, whose implementations follow the official MMAUD repository where available.\footnote{\url{https://github.com/ntu-aris/MMAUD}} 
Our method achieves a 3D error of 0.30~m, only 0.01~m above the best audio-visual method \cite{LEI_ESWA}, while obtaining the best $P_{95}$, $SR_{0.3}$, and $SR_{0.5}$. More importantly, the adaptive controller selects an average window of only 0.23~s, reducing the temporal correspondence latency to 0.136~s, substantially below all fixed-window baselines. Although several audio-visual methods benefit from complementary visual information and achieve strong localization accuracy, our adaptive audio localization model remains highly competitive, while providing a substantially better accuracy-latency trade-off.

\begin{figure}
\centering
\includegraphics[width=2.7in]{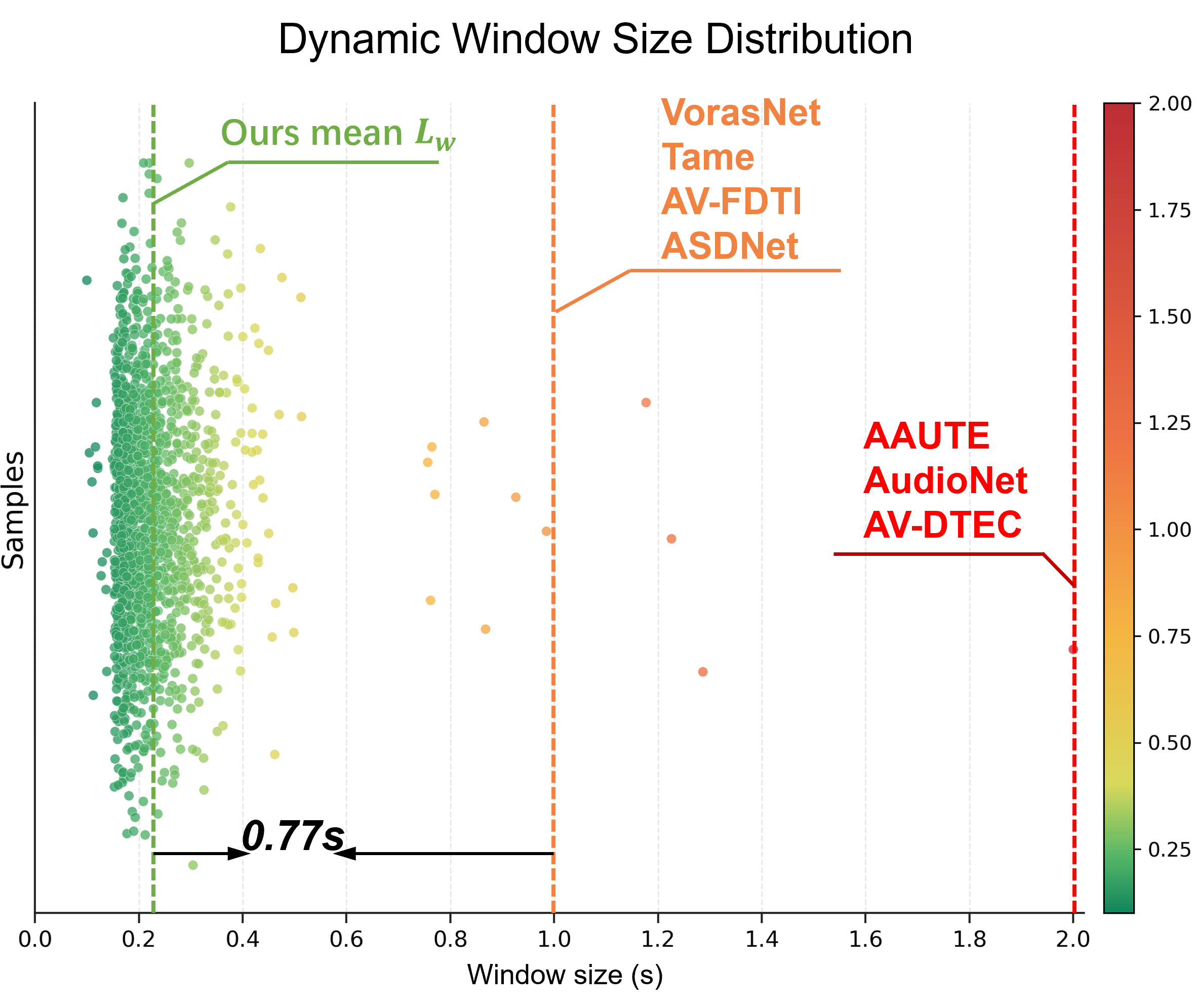}
\caption{The visualization of the dynamic window size distribution and comparison with SOTA methods.}
\label{fig:window}
\end{figure}

Fig.~\ref{fig:traj_comparison} further shows that long-window methods exhibit a clear motion-direction offset at $T=187$~s, whereas our adaptive window yields an estimate closer to the current ground truth. Fig.~\ref{fig:window} confirms that dynamically selected windows are substantially shorter than the fixed windows used by existing methods.

\textbf{Cross-domain robustness:} All methods are trained on MMAUD V1 and zero-shot transferred to the more challenging MMAUD V2 scenes, as shown in Fig.~\ref{fig:cross}, which contain denser noise and more distant UAV targets. As shown in Table~\ref{table:comparisonmmaud2_cross}, our controller selects longer windows to accumulate sufficient acoustic cues. Despite the advantage of visual information, our method outperforms all audio-visual baselines while achieving the lowest temporal correspondence latency.

\begin{table}
\centering
\caption{Ablation study of \textbf{fixed audio window} sizes on MMAUD-V1~\cite{MMAUD}, evaluated by the 3D position error $E$ (m). \textbf{\textit{Bold italic}} values indicate the results obtained using the audio segment lengths adopted in the original papers.}
\label{table:fixedwindowcompare}
\renewcommand{\arraystretch}{1.1}
\setlength{\tabcolsep}{9.0pt}
\small
\begin{tabular}{lccccc}
\toprule
\textbf{Methods} 
& \multicolumn{1}{c}{\textbf{0.1s}} 
& \multicolumn{1}{c}{\textbf{0.23s}} 
& \multicolumn{1}{c}{\textbf{0.5s}} 
& \multicolumn{1}{c}{\textbf{1.0s}}
& \multicolumn{1}{c}{\textbf{2.0s}} 
\\ 
\hline
\textbf{AudioNet} \cite{yang2023av}  & 3.78 & 1.34 & 0.84 & 0.69 & \textbf{\textit{0.76}} \\
\textbf{VorasNet} \cite{vora2023dronechase} & 2.62 & 1.89 & 1.17 & \textbf{\textit{1.06}} & 0.98 \\
\textbf{TAME} \cite{tame} & 1.42 & 0.89 & 0.53 & \textbf{\textit{0.55}} & 0.57 \\
\textbf{AAUTE} \cite{AAUTE} & 3.94 & 2.43 & 1.12 & 0.89 & \textbf{\textit{0.48}} \\
\hline
\textbf{ASDNet} \cite{tao2021someone} & 2.45 & 1.05 & 0.85 & \textbf{\textit{0.99}} & 0.89 \\
\textbf{AV-PED} \cite{yang2023av} & 1.99 & 1.23 & 0.83 & \textbf{\textit{0.87}} & 0.78 \\
\textbf{AV-FDTI} \cite{YANG2024} & 2.45 & 1.56 & 0.89 & \textbf{\textit{0.53}} & 0.43\\
\textbf{Lei et al.} \cite{LEI_ESWA} & 1.21 & 0.78 & 0.35 & \textbf{\textit{0.29}} & 0.43 \\
\textbf{AV-DTEC} \cite{av-dtec} & 2.82 & 1.62 & 0.98 & 0.76 & \textbf{\textit{0.58}} \\
\hline
\textbf{Ours-Fixed} & 1.02 & \textbf{\textit{0.63}} & 0.79 & 0.98 & 1.29 \\
\textbf{Ours-Adaptive} & - & \textbf{\textit{0.30}} & - & - & - \\
\bottomrule
\end{tabular}
\end{table}

\vspace{-1em}
\subsection{Ablation Study}

\textbf{Ablation on Fixed Window Size:}
Each competing method is retrained for every fixed window in Table III. Ours-Fixed disables PPO and adaptive feature modulation. Ours-Adaptive uses variable windows averaging 0.23s. The complete adaptive framework reduces error from 0.63m to 0.30m at the same mean window length.
Also, the original baseline windows are not always optimal: TAME\cite{tame}, ASDNet\cite{tao2021someone}, and AV-PED\cite{yang2023av} achieve lower errors at 0.5s than at 1.0s, with half the window-induced latency.

\vspace{1em}

\vspace{-1em}
\textbf{Ablation on State composition: }
As shown in Table~\ref{table:ablationstate}, using $Q_k$ alone favors shorter windows but degrades localization accuracy, since a dominant GCC peak may still be unreliable. Adding $D_k$ helps identify temporally unstable TDOAs and retain longer windows when necessary, substantially reducing the error. Incorporating $C_k^{\mathrm{clo}}$ further enforces cross-pair spatial consistency, enabling shorter yet more reliable observations. The temporal cue $\bar{\xi}_{k-1}$ provides tracking feedback and further improves the accuracy--latency trade-off. Finally, feature control further reduces the localization error under the same average window length.

\begin{table}
\centering
\caption{Ablation study on the \textbf{state composition} and the \textbf{feature control} (FC) of the PPO-based adaptive controller.}
\label{table:ablationstate}
\renewcommand{\arraystretch}{1.0}
\setlength{\tabcolsep}{9.0pt}
\small
\begin{tabular}{ccccccc}
\toprule
\textbf{$Q_k$}
& \textbf{$D_k$}
& \textbf{$C_k^{\mathrm{clo}}$}
& \textbf{$\bar{\xi}_{k-1}$}
& \textbf{FC}
& \textbf{$L_w\downarrow$}
& \textbf{$E\downarrow$}
\\
\midrule
 & & & & &  2.00 & 1.29 \\
\checkmark & & & & & 0.58 & 1.38 \\
\checkmark & \checkmark & & & & 0.92 & 0.81 \\
\checkmark & \checkmark & \checkmark & & & 0.46 & 0.55 \\
\checkmark & \checkmark & & \checkmark & & 0.41 & 0.53 \\
\checkmark &  & \checkmark & \checkmark & & 0.32 & 0.61 \\
\midrule
\textbf{\checkmark}
& \textbf{\checkmark}
& \textbf{\checkmark}
& \textbf{\checkmark}
&
& \textbf{0.23}
& 0.39 \\

\textbf{\checkmark}
& \textbf{\checkmark}
& \textbf{\checkmark}
& \textbf{\checkmark}
& \textbf{\checkmark}
& \textbf{0.23}
& \textbf{0.30} \\

\bottomrule
\end{tabular}
\end{table}






\begin{table}
\centering
\caption{Ablation study on \textbf{different control policy} designs. 
$L_w$ denotes the average audio window length in seconds.}
\label{table:ablationpolicy}
\renewcommand{\arraystretch}{1.1}
\setlength{\tabcolsep}{5.0pt}
\small
\begin{tabular}{lcccccc}
\toprule
\textbf{Policy}
& \textbf{$D_x\downarrow$}
& \textbf{$D_y\downarrow$}
& \textbf{$D_z\downarrow$}
& \textbf{$E\downarrow$}
& \textbf{$L_{w}\downarrow$}
& \textbf{$L_{t}\downarrow$}
\\
\midrule

Fixed
& 0.32
& 0.49
& 0.36
& 0.63
& \textbf{0.23}
& \textbf{0.132}
\\

Grid Search
& 0.27
& 0.35
& 0.28
& 0.54
& \underline{0.28}
& 0.164
\\

Heuristic
& 0.30
& 0.43
& 0.35
& 0.64
& 0.36
& 0.208
\\

MLP imitation
& 0.23
& 0.31
& 0.25
& 0.46
& 0.31
& 0.182
\\

\midrule

\textbf{Ours-PPO}
& \underline{0.17}
& \underline{0.18}
& \textbf{0.12}
& \underline{0.30}
& \textbf{0.23}
& \underline{0.136}
\\

Oracle
& \textbf{0.15}
& \textbf{0.15}
& \underline{0.18}
& \textbf{0.28}
& 0.33
& 0.178
\\

\bottomrule
\end{tabular}
\end{table}


\textbf{Ablation on Adaptive Policy:}
Table~\ref{table:ablationpolicy} compares Ours-PPO with fixed, grid-search, heuristic, imitation, and oracle policies. The fixed policy uses a constant 0.23s window, while grid search selects the globally best discrete window--feature configuration on the training set. The heuristic adaptively adjusts both controls using hand-crafted rules, whereas the MLP imitates oracle actions from the training set. The oracle selects the minimum-error candidate at each test step using ground truth and is therefore non-deployable. PPO achieves 0.30m error with a mean window of 0.23s, outperforming all deployable alternatives and approaching the 0.28m oracle bound without test-time labels.

\textbf{Centered vs. Causal Window: }
We further compare centered and causal windows on MMAUD using the same localization architecture, where each causal segment is aligned with the GT at its end timestamp. With fixed causal windows of 0.23s, 1.0s, and 2.0s, the localization errors are 3.84m, 2.87m, and 3.02m, respectively. Although causal windows eliminate the window-induced correspondence latency and reduce the latency to $L_i$, they result in substantially larger errors than centered windows, indicating the importance of temporal alignment for accurate localization of moving UAVs.

\section{Conclusion}

We presented a closed-loop audio-based UAV localization
framework that uses acoustic consistency cues and PPO
to jointly adapt window length and Mamba temporal
feature contribution.
Experiments on MMAUD demonstrate competitive localization
accuracy with reduced temporal correspondence latency
and improved cross-domain performance relative to
the evaluated baselines.

\section*{ACKNOWLEDGMENT}
The authors acknowledge the use of large language models (e.g., ChatGPT by OpenAI) for assisting in prototype code generation, partial methodology modules, and improving the clarity of the manuscript. All technical contributions and interpretations remain the responsibility of the authors.

\bibliographystyle{IEEEtran}
\hypersetup{urlcolor=black}

\bibliography{mybib}

\end{document}